\documentclass{article}

\usepackage{arxiv}

\usepackage[utf8]{inputenc} 
\usepackage[T1]{fontenc}    
\usepackage{hyperref}       
\usepackage{url}            
\usepackage{booktabs}       
\usepackage{amsfonts}       
\usepackage{nicefrac}       
\usepackage{microtype}      
\usepackage{lipsum}		
\usepackage{graphicx}
\usepackage{natbib}
\usepackage{doi}
\usepackage{xcolor}

\title{Recipro-CAM: Fast Gradient-Free Visual  Explanations for Convolutional Neural Networks}

\author{ 
Seok-Yong~Byun \\
Intel Corp. \\
Seoul, South Korea 06134 \\
\texttt{mark.byun@intel.com}  \\
\And
Wonju~Lee \\
Intel Corp.\\
Seoul, South Korea 06134 \\
\texttt{wonju.lee@intel.com} \\
}

\renewcommand{\shorttitle}{\textit{arXiv}}

\begin{document}
\maketitle

\begin{abstract}
The Convolutional Neural Network (CNN) is a widely used deep learning architecture for computer vision. However, its black box nature makes it difficult to interpret the behavior of the model. To mitigate this issue, AI practitioners have explored explainable AI methods like Class Activation Map (CAM) and Grad-CAM. Although these methods have shown promise, they are limited by architectural constraints or the burden of gradient computing.
To overcome this issue, Score-CAM and Ablation-CAM have been proposed as gradient-free methods, but they have longer execution times compared to CAM or Grad-CAM based methods, making them unsuitable for real-world solution though they resolved gradient related issues and enabled inference mode XAI.
To address this challenge, we propose a fast gradient-free Reciprocal CAM (Recipro-CAM) method. Our approach involves spatially masking the extracted feature maps to exploit the correlation between activation maps and network predictions for target classes. Our proposed method has yielded promising results, outperforming current state-of-the-art method in the Average Drop-Coherence-Complexity (ADCC) metric by $1.78 \%$ to $3.72 \%$, excluding VGG-16 backbone.
Moreover, Recipro-CAM generates saliency maps at a similar rate to Grad-CAM and is approximately $148$ times faster than Score-CAM.
The source code for Recipro-CAM is available in our data analysis framework.\footnote{https://github.com/openvinotoolkit/datumaro/blob/feats/recipro-cam/datumaro/components/algorithms/recipro\_cam.py}.
\end{abstract}

\keywords{Computer Vision \and Class Activation Map \and Explainable AI \and Gradient-free CAM \and White-Box XAI}

\section{Introduction}

Convolutional Neural Networks (CNNs) are a popular choice for computer vision tasks due to their high accuracy and cost-effectiveness. However, they can fail to work as expected for edge cases or out-of-distribution inputs, which is a significant problem in critical applications such as medical diagnosis or autonomous driving. The lack of transparency and interpretability in the model's decision-making process has led to the emergence of Explainable AI (XAI) as a solution. This can help practitioners identify and address potential biases, errors, or limitations in the model's design or training data.


The class activation map (CAM) method (\cite{Zhou2016}) is a simple and effective approach to generate saliency maps that indicate the contribution of different image regions to the model's output. However, CAM has some architectural requirements that limit its applicability to certain models. To address this issue, Grad-CAM (\cite{Selvaraju2016}) was proposed, which uses gradient information to generate saliency maps and can work with any CNN architecture. 

Recently, gradient-free methods for inferring saliency maps have been proposed, such as Score-CAM and Ablation-CAM by ~(\cite{Wang2020}) and ~\cite{Desai2020}, respectively. Unlike Grad-CAM, Score-CAM does not require gradient computations, which eliminates any gradient-related restrictions. However, Score-CAM's execution time depends on input resolution, convolution channel dimensions, and network capacity, making it slower than CAM or Grad-CAM methods by approximately $127 \times$. This motivated us to develop a fast gradient-free XAI solution.

\begin{figure}[t]
  \begin{center}
    \includegraphics[width=3.4in]{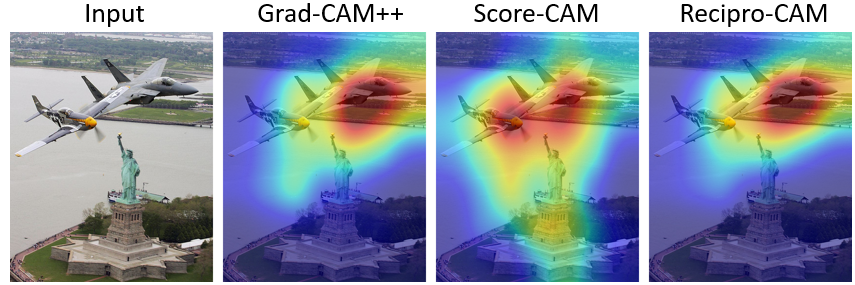}
  \end{center}
  \caption{Comparing the resolution capabilities of Grad-CAM++, Score-CAM, and Recipro-CAM, it was found that while the predicted class was an aircraft carrier, the ground truth was a warplane. However, Score-CAM was unable to distinguish the warplane from a nearby object.}
  \label{fig:main}
\end{figure}

Recipro-CAM draws inspiration from two methods, CAM and RISE ~(\cite{Petsiuk2018}), which establish a reciprocal relationship between the convolution layer's activation map and the network's output. Unlike CAM, which uses only the activation map and the last FC layer's weight, or Score-CAM, which generates new input images from the convolution activation map, Recipro-CAM focuses on the reciprocal relationship between the convolution layer's activation map and the output prediction itself. This approach is illustrated in Figure~\ref{fig:sm}. A detailed explanation of the Recipro-CAM method will be provided in the following section~\ref{sec:prop}. 

The main contributions of this paper include:

\begin{itemize}
\item We present a new, efficient, and gradient-free white-box XAI solution that utilizes the reciprocal relationship between the convolution layer's activation map and the network's prediction result. Our method, Recipro-CAM, removes both architectural and gradient-related constraints, similar to Score-CAM, while providing approximately $148 \times$ faster execution performance than Score-CAM.
\item We evaluate the classification accuracy of Recipro-CAM using several metrics, including Average Drop, Average Increase, Deletion, Insertion, Coherency, Complexity, and Average DCC (ADCC), with VGG-16, ResNet, and ResNeXt architectures. Our results demonstrate state-of-the-art accuracy on the ADCC metric for all architectures, except for VGG-16. Additionally, our method achieves relatively high accuracy values for other $4$ metrics compared to previous methods.
\end{itemize}

\section{Related Work}
Deep learning is a perplexing field due to its black-box nature. The intricate internal mechanisms and reactions that occur between input and output are hidden, which makes it challenging for AI practitioners to comprehend neuron reactions to input variations. However,\cite{Zeiler2014} and \cite{Springenberg2015} have proposed a deconvolution approach that uncovers each layer's learned features and important pixels in the input image in a class-agnostic way. These breakthroughs have contributed to the development of CNN architectures and helped to explain CNN's layer-wise learning property. Yet, understanding these features alone is insufficient to explain the prediction results for specific inputs. As a result, researchers have sought out ways to clarify the relationship between input data and CNN output.

The first solution suggested to address this issue is CAM~\cite{Zhou2016}. This method produces a map that highlights the important regions of an image for a particular class by multiplying a global average pooling activation vector with a fully connected weight vector specific to the class. Essentially, the saliency map $S^c$ for a given class $c$ is obtained by
\begin{equation}
S^{c} = \sum_{k} w_{k,c}\sum_{u,v} f_{k}(u,v)
\end{equation}
where $w_{k,c}$ is the last FC layer’s weight between channel $k$ and class $c$ and $f_{k}(u,v)$ is the activation at $(u,v)$ of channel $k$.
CAM allows AI practitioners not only to analyze the capacity of their neural network architecture but also to understand how the network reacts to specific classes of input data. However, this method has a limitation in that it requires the presence of a global average or max pooling layer in the architecture. This means that certain neural network architectures may not be compatible with CAM method.

Grad-CAM~(\cite{Selvaraju2016}), Grad-CAM++~(\cite{Chattopadhay2018}), Axiom-based Grad-CAM~(\cite{Fu2020}), and Smooth Grad-CAM++~(\cite{Omeiza2019}) are methods for generating saliency maps that address limitations of CAM method and can be applied more broadly. These methods rely on the gradient information of the class confidence output with respect to the activation map of the convolution layer as
\begin{equation}
\mathcal{L}_{\textrm{Grad-GAM}}^{c} = \textrm{ReLU}\left(\sum_{k}\alpha_{k}^{c}A^{k}\right),
\end{equation}
where 
\begin{equation}
\alpha_{k}^{c} = \frac{1}{Z}\sum_i\sum_j\frac{\partial y^{c}}{\partial A^{k}_{ij}}.
\end{equation}
Gradient-based approaches have emerged as a popular solution, as they can overcome the limitations of CAM and provide interpretability. However, these methods require a trainable model to utilize gradient information, which restricts their use in post-deployment frameworks like ONNX~(\cite{onnx2019}) or OpenVINO~(\cite{openvino2019}). Moreover, researchers have identified issues such as saturation and false confidence with these methods~\cite{Wang2020}, leading to the development of Score-CAM~(\cite{Wang2020}), Smooth Score-CAM~(\cite{Wang2021}), and Integrated Score-CAM~(\cite{Naidu2020}), which have shown improved performance.

Score-CAM is a technique that extracts activation information from convolution layers and uses it to generate a saliency map that highlights the important regions of an input image for a specific class. This is done by first masking the input image with the channel-wise activation map, which is then used to calculate the channel-wise increase of confidence (CIC) for the specified class. Unlike other techniques such as CAM or Grad-CAM, Score-CAM does not rely on gradient information and provides improved explainability. However, generating the saliency map using Score-CAM requires multiple inference operations, which increases the computation time compared to other approaches. The number of operations also depends on the number of channels in the convolution layer.

The authors of \cite{Desai2020} have also proposed a novel gradient-free approach named by Ablation-CAM. This method overcomes the saturation and gradient explosion problems of Grad-CAM by introducing the effective slope, which is calculated as the difference between the original prediction score and the prediction score from the ablated activation map. While the ablation drop ~(\cite{Morcos2018}) is helpful in describing how important an ablation unit is for a target class, it requires significant time due to the need to iterate over each feature map to ablate.

In addition to white-box approaches, there have been several studies on black-box approaches, such as ~\cite{Petsiuk2018,Kenny2021,Dabkowski2017,Petsiuk2020}. 
These approaches can generate a saliency map without relying on network architectural information or gradient computability. However, they typically require much more time than white-box methods and may exhibit relatively lower accuracy.

\subsection{Evaluation Metrics}
The effectiveness of XAI solutions should be measured using standardized metrics in an unified manner. However, until \cite{Poppi2021} proposed ADCC metric, there was no universally accepted standard metric. Previous studies, such as ~\cite{Chattopadhay2018,Petsiuk2018,Fu2020}, had suggested their own metrics such as Avg Drop, Avg Inc, Deletion, and Insertion, however these standalone metrics could lead incorrect performance evaluations, as demonstrated by \cite{Poppi2021} using Fake-CAM. 

ADCC is a comprehensive metric that takes into account the average drop, coherence, and complexity, and provides a single score as their harmonic mean with following equation
\begin{equation} \label{eq:adcc}
\textrm{ADCC}(x)= 3\Bigg(\frac{1}{\textrm{Coherency}(x)}+\frac{1}{1-\textrm{Complexity}(x)} +\frac{1}{1-\textrm{AverageDrop}(x)}\Bigg)^{-1}.
\end{equation}
We will evaluate the precision of our approach by using ADCC metric.

\begin{figure*}[t]
  \begin{center}
    \includegraphics[width=6in]{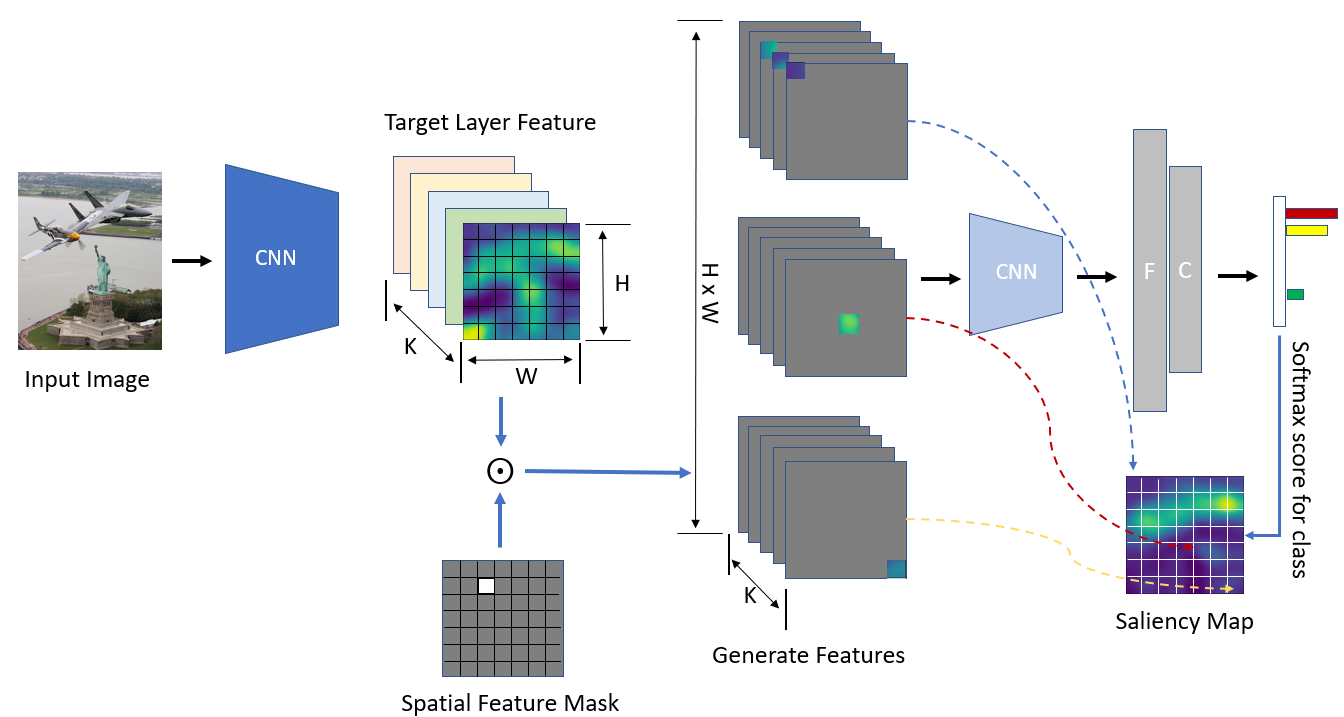}
  \end{center}
  \caption{Recipro-CAM works as follows: The target layer's feature map is multiplied by a set of $HW$ spatial masks, where each mask has a single value at a specific spatial position in the feature map. This generates a new set of $HW$ feature maps, which are then fed into the next part of the network. The predicted scores for the target class are collected and used to fill each position of the saliency map. Here, $K$ represents the number of channels, $H$ represents the feature height, and $W$ represents the feature width.}
  \label{fig:sm}
\end{figure*}

\section{Recipro-CAM}\label{sec:prop}
The proposed Recipro-CAM is a lightweight saliency map generation method that does not rely on gradient or architecture information. As shown in Figure ~\ref{fig:sm}, Recipro-CAM extracts a $K \times H \times W$ dimensional feature map $F$ from a specified convolution layer and generates a set of $H \times W$ spatial masks. 
These masks are used to generate new input feature maps that are passed through the rest of the architecture, allowing the proposed method to compute prediction scores for each class at every feature location $(u,v)$ imposed by the masks. For simplicity, we will ignore the batch dimension in this paper.


\subsection{Spatial Mask and Perturbed Feature Map Generation}
\begin{figure}[t]
  \begin{center}
    \includegraphics[width=3.2in]{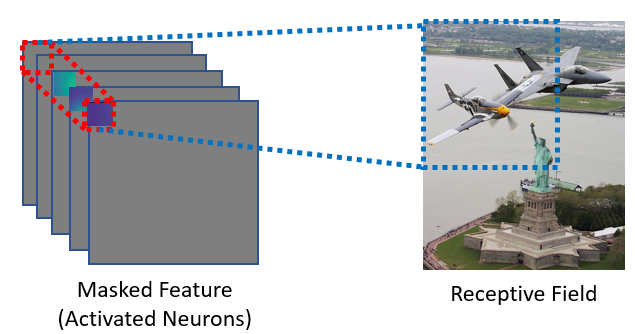}
  \end{center}
  \vspace{-5mm}
  \caption{Samples of receptive fields covered by enabled neurons are marked with dashed red boxes.}
  \label{fig:rf}
  \vspace{-2mm}
\end{figure}

To emphasize the significance of each activation on the feature map in relation to the output prediction, we only keep the activation at the target pixel, which corresponds to the specific region in the original input image based on the receptive field. This is illustrated in Figure~\ref{fig:rf}. We assume that each perturbed feature map produced by the spatial mask can capture a wide range of characteristics of the input image. Therefore, a feature map with multiple channels enabled at a single position can provide enough information for prediction with less overlap of information from nearby positions.

The process of generating a spatial mask $M^n$ involves selecting one pixel in the feature map setting it as $1$, while setting all other pixels as $0$. This is done for every pixel in the feature map to create $N$ spatial masks, where $N$ equals the product of the feature map's height and width, i.e., $N=H\times W$. Each mask corresponds to a unique pixel position in the feature map.

The masked feature map $\tilde{F}$ is created by applying each spatial mask $M^n$ to the original feature map $F$ extracted from the convolution layer. This is done through element-wise multiplication of the feature map with the spatial mask $M$. The resulting masked feature map $\tilde{F}$ highlights the activations of the specific pixel position in the feature map that the spatial mask corresponds to, while setting all other activations to zero.

By generating $N$ different masked feature maps, we obtain $N$ different views of the original feature map, each highlighting a different pixel position. This allows us to analyze the activations of the feature map at a more fine-grained level and understand which pixels contribute the most to the final output. 
The $n$th masked feature map for channel $k$ can be expressed as
\begin{equation}
\tilde{F}_{k}^{n} = F_{k} \odot M^{n}
\end{equation}
with the Hardamard product $\odot$.


\subsection{Saliency Map Generation}
Recipro-CAM generates a saliency map by dividing the network into two parts based on the specified convolution layer. The first part, denoted as $f$, represents the feature generation network, while the second part, denoted as $g$, represents the subsequent network layers.
Then a saliency map $S^c$ for a class $c$ can be expressed as 
\begin{equation}
S^c = \textrm{reshape}\left[\frac{\mathbf{y}_c - \min(\mathbf{y}_c)}{\max(\mathbf{y}_c)-\min(\mathbf{y}_c)}, (H,W) \right],
\end{equation}
where the $N\times 1$ prediction vector $\mathbf{y}_c=\left[y_{c}^{1}, \dots, y_{c}^{N}\right]^{T}$ for a class $c$ is composed of
\begin{equation}
y_{c}^{n} = \textrm{softmax}\left(g(f(I)\odot M^{n})\right)_{c}
\end{equation}
for $n=1,\dots,N$.

\section{Experiments}
In Sections~\ref{sec:quant},~\ref{sec:qual}, and~\ref{sec:perf}, we will provide a quantitative, qualitative, and performance analysis, respectively, to demonstrate the accuracy and effectiveness of Recipro-CAM. 

\subsection{Quantitative Analysis} \label{sec:quant}
For quantitative analysis, we adopted the ADCC metric proposed by~\cite{Poppi2021} and followed their experimental conditions. We evaluated the performance of Recipro-CAM on the ILSVRC2012~(\cite{Russakovsky2015}) validation set, which contains $50,000$ images. To compare our results with those of ~\cite{Poppi2021}, we also evaluated Recipro-CAM on several models, including VGG-16~(\cite{Simonyan2015}), ResNet-18, ResNet-50, ResNet-101~(\cite{He2016}), ResNeXt-50, and ResNeXt-101~(\cite{Xie2017}). We applied each CAM approach on the last block or convolution layer of these models. Prior to evaluation, we resized the input images to $256 \times 256$ and center cropped them to $224 \times 224$. The images were normalized using a mean of $[0.485, 0.456, 0.406]$ and a standard deviation of $[0.229, 0.224, 0.225]$. Finally, we added the results obtained by Recipro-CAM to those obtained by~\cite{Poppi2021}, and summarized the consolidated results in Table~\ref{tb:perf}.

Table~\ref{tb:perf} presents the performance comparison of Recipro-CAM with other CAM approaches on the ADCC metric. The results show that Recipro-CAM achieves state-of-the-art performance on the ADCC metric for five architectures, except for VGG-16. Upon closer inspection of the ADCC metrics, Score-CAM was found to provide the highest Average Drop score for the five architectures, while Recipro-CAM was significantly less complex, leading to superior performance in the ADCC metric, which combines Average Drop, Coherency, and Complexity. However, for VGG-16, the opposite trend was observed. In terms of Coherency, Score-CAM appears to be generally better than the proposed method, but there is no significant difference to affect the ADCC ranking.

Regarding the Insertion metric, Recipro-CAM and Score-CAM show alternating first and second places. The Deletion metric usually displays results similar to those of the Insertion metric, but in ResNext model, Score-CAM outperforms the proposed method. However, it is worth noting that ~\cite{Poppi2021} have shown that the Average Drop, Average Increase, Insertion, and Deletion metrics are insufficient for evaluating the explainability of models using Fake-CAM. Instead, they proposed the ADCC metric as a more reasonable and separable evaluation method. We also observed this in our results.

By default, our mask generation method employs a Dirac delta kernel for all experiments. However, other kernel functions, such as Gaussian, can also be used for mask generation. We evaluated the ADCC scores for ResNet-50 and 101 models using a $3 \times 3$ Gaussian kernel in Table~\ref{tb:kernel} and found that the scores did not differ significantly from those obtained using the default method with a Dirac delta kernel. This indicates that a channel vector created by a single-pixel mask (as shown in Figure~\ref{fig:rf}) can accurately represent saliency information for the receptive area of the input image.

\begin{table}[]
\caption{Comparison of different CAM-based approaches using existing metrics on six different backbones. The evaluation scores for other CAM methods were obtained from~\cite{Poppi2021}.}
\scriptsize \centering
\begin{tabular}{lccccccc|ccccccc} \hline
& \multicolumn{7}{c|}{VGG-16} & \multicolumn{7}{c}{ResNet-18} \\ \hline
Method & \begin{tabular}[c]{@{}c@{}} \textbf{Drop} \\ $(\downarrow)$ \end{tabular} & \begin{tabular}[c]{@{}c@{}}Inc \\ $(\uparrow)$ \end{tabular} & \begin{tabular}[c]{@{}c@{}}Del \\ $(\downarrow)$ \end{tabular} & \begin{tabular}[c]{@{}c@{}}Ins \\ $(\uparrow)$ \end{tabular} & \begin{tabular}[c]{@{}c@{}} \textbf{Coher} \\ $(\uparrow)$ \end{tabular} & \begin{tabular}[c]{@{}c@{}} \textbf{Compl} \\ $(\downarrow)$ \end{tabular} & \begin{tabular}[c]{@{}c@{}} \textbf{ADCC} \\ $(\uparrow)$ \end{tabular} & \begin{tabular}[c]{@{}c@{}} \textbf{Drop} \\ $(\downarrow)$ \end{tabular} & \begin{tabular}[c]{@{}c@{}}Inc \\ $(\uparrow)$ \end{tabular} & \begin{tabular}[c]{@{}c@{}}Del \\ $(\downarrow)$ \end{tabular} & \begin{tabular}[c]{@{}c@{}}Ins \\ $(\uparrow)$ \end{tabular} & \begin{tabular}[c]{@{}c@{}} \textbf{Coher} \\ $(\uparrow)$ \end{tabular} & \begin{tabular}[c]{@{}c@{}} \textbf{Compl} \\ $(\downarrow)$ \end{tabular} & \begin{tabular}[c]{@{}c@{}} \textbf{ADCC} \\ $(\uparrow)$ \end{tabular} \\ \hline
Grad-CAM & 66.42 & 5.92 & 11.12 & 19.56 & 69.20 & 15.65 & 53.52 & 42.90 & 16.63 & 13.43 & 41.47 & 81.03 & 23.04 & 69.98 \\
Grad-CAM++ & 32.88 & 20.10 & 8.82 & 36.60 & 89.34 & 26.33 & 75.65 & 17.85 & 34.46 & 12.30 & 44.80 & 98.18  & 44.63 & 74.24 \\
SGrad-CAM++ & 36.72 & 16.11 & 10.57 & 31.36 & 82.68 & 28.09 & 71.72 & 20.67 & 29.99 & 12.83 & 43.13 & 97.53 & 43.11 & 74.20 \\
Score-CAM & 26.13 & 24.75 & 9.52 & 47.00 & 93.83 & 20.27 & \textbf{81.66} & 12.81 & 40.41 & 10.76 & 46.01 & 98.35 & 41.78 & 77.30 \\
\textbf{Recipro-CAM} & 21.51 & 34.86 & 9.50 & 46.88 & 92.24 & 27.48 & \textcolor{blue}{80.27} & 20.68 & 36.30 & 10.19 & 44.93 & 97.38 & 33.60 & \textbf{79.08} \\ \hline
& \multicolumn{7}{c|}{ResNet-50} & \multicolumn{7}{c}{ResNet-101} \\ \hline
Grad-CAM & 32.99 & 24.27 & 17.49 & 48.48 & 82.80 & 22.24 & 75.27 & 29.38 & 29.35 & 18.66 & 47.47 & 81.97 & 22.51 & 76.40 \\
Grad-CAM++ & 12.82 & 40.63 & 14.10 & 53.51 & 97.84 & 43.99 & 75.86 & 11.38 & 42.07 & 14.99 & 56.65 & 98.28 & 43.94 & 76.34 \\
SGrad-CAM++ & 15.21 & 35.62 & 15.21 & 52.43 & 97.47 & 42.25 & 76.19 & 13.37 & 37.76 & 14.32 & 58.23 & 97.76 & 42.61 & 76.54 \\
Score-CAM & 8.61 & 46.00 & 13.33 & 54.16 & 98.12 & 42.05 & 78.14 & 7.20 & 47.93 & 14.63 & 59.57 & 98.37 & 42.04 & 78.55 \\
\textbf{Recipro-CAM} & 15.69 & 40.54 & 13.34 & 55.39 & 96.68 & 32.90 & \textbf{80.84} & 15.07 & 41.39 & 15.80 & 59.28 & 97.21 & 32.45 & \textbf{81.38} \\ \hline
& \multicolumn{7}{c|}{ResNeXt-50} & \multicolumn{7}{c}{ResNeXt-101} \\ \hline
Grad-CAM & 28.06 & 29.42 & 20.73 & 50.30 & 82.72 & 25.57 & 76.09 & 24.12 & 36.37 & 20.47 & 61.04 & 82.94& 25.45 & 77.62 \\
Grad-CAM++ & 11.12 & 41.38 & 17.07 & 56.06 & 97.30 & 48.66 & 73.16 & 9.74 & 42.63 & 17.63 & 62.90 & 95.05 & 46.27 & 74.61 \\
SGrad-CAM++ & 12.70 & 36.58 & 16.90 & 56.76 & 97.32 & 47.48 & 73.58 & 9.49 & 40.43 & 17.67 & 64.16 & 96.81 & 49.24 & 73.03 \\
Score-CAM & 7.20 & 45.70 & 15.59 & 57.92 & 98.00 & 46.86 & 75.38 & 5.37 & 47.70 & 17.30 & 63.61 & 97.03 & 46.83 & 75.60 \\
\textbf{Recipro-CAM} & 13.70 & 40.82 & 18.94 & 58.93 & 96.37 & 37.36 & \textbf{79.10} & 12.03 & 42.69 & 20.25 & 64.70 & 97.50 & 35.62 & \textbf{80.74} \\ \hline
\end{tabular} \label{tb:perf}
\vspace{-3mm}
\end{table}

\begin{table}[]
\caption{Comparison of different mask generation methods.}
\scriptsize \centering
\begin{tabular}{lcccc|cccc} \hline
& \multicolumn{4}{c|}{ResNet-50} & \multicolumn{4}{c}{ResNet-101} \\ \hline
Method & \begin{tabular}[c]{@{}c@{}}Drop \\ $(\downarrow)$ \end{tabular} & \begin{tabular}[c]{@{}c@{}}Coher \\ $(\uparrow)$ \end{tabular} & \begin{tabular}[c]{@{}c@{}}Compl \\ $(\downarrow)$ \end{tabular} & \begin{tabular}[c]{@{}c@{}} \textbf{ADCC} \\ $(\uparrow)$ \end{tabular} & \begin{tabular}[c]{@{}c@{}}Drop \\ $(\downarrow)$ \end{tabular} & \begin{tabular}[c]{@{}c@{}}Coher \\ $(\uparrow)$ \end{tabular} & \begin{tabular}[c]{@{}c@{}}Compl \\ $(\downarrow)$ \end{tabular} & \begin{tabular}[c]{@{}c@{}} \textbf{ADCC} \\ $(\uparrow)$ \end{tabular} \\ \hline
Recipro-CAM (Default) & 15.69 & 96.68 & 32.90 & \textbf{80.84} & 15.07 & 97.21 & 32.45 & \textbf{81.38} \\
Recipro-CAM (Gaussian) & 18.69 & 96.56 & 30.53 & \textbf{80.97} & 17.87 & 97.16 & 30.18 & \textbf{81.54} \\ \hline
\end{tabular} \label{tb:kernel}
\vspace{-3mm}
\end{table}

\subsection{Qualitative Analysis}\label{sec:qual}
For our qualitative evaluation, we used ResNet-50 backbone architecture and the Torchvision ImageNet pretrained model. To generate saliency maps, we utilized the torchCAM~(\cite{Fernandez2020}) library, and applied the same pre-processing and normalization methods discussed in Section~\ref{sec:quant} to the input images.

To conduct a systematic analysis, we categorized the input images from the ILSVRC2012 validation dataset into three groups based on the number and type of objects present. The first group contained single object-only cases, the second group had multiple objects of the same type, and the third group had multiple objects of different types.

\begin{figure}[t]
\begin{center}
\includegraphics[width=6.5in]{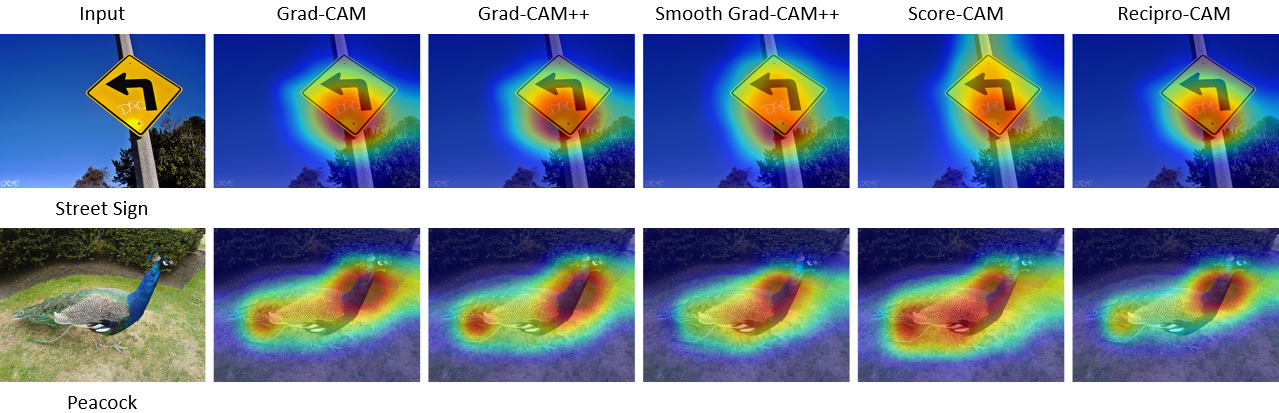}
\end{center}
\vspace{-3mm}
\caption{Single-object saliency map results for Street Sign and Peacock inputs processed with Grad-CAM, Grad-CAM++, Smooth Grad-CAM++, Score-CAM, and Recipro-CAM to generate saliency maps.}
\label{fig:ex1}
\vspace{-2mm}
\end{figure}

In Figure~\ref{fig:ex1}, we present the results of the first group. For the street sign image, all methods focused on the bottom edge and nail head. However, Score-CAM showed additional saliency on the upper pole part and treetops, while Recipro-CAM exhibited a relatively small saliency area for the treetops.
For the peacock image, Recipro-CAM showed an overall similar saliency pattern to Grad-CAM and Grad-CAM++, except for the tail part. In contrast, Score-CAM's saliency map covered the entire peacock, and Smooth Grad-CAM++ focused solely on the body part. Therefore, in this case, Score-CAM provided a more explainable result.

\begin{figure}[t]
\begin{center}
\includegraphics[width=6.5in]{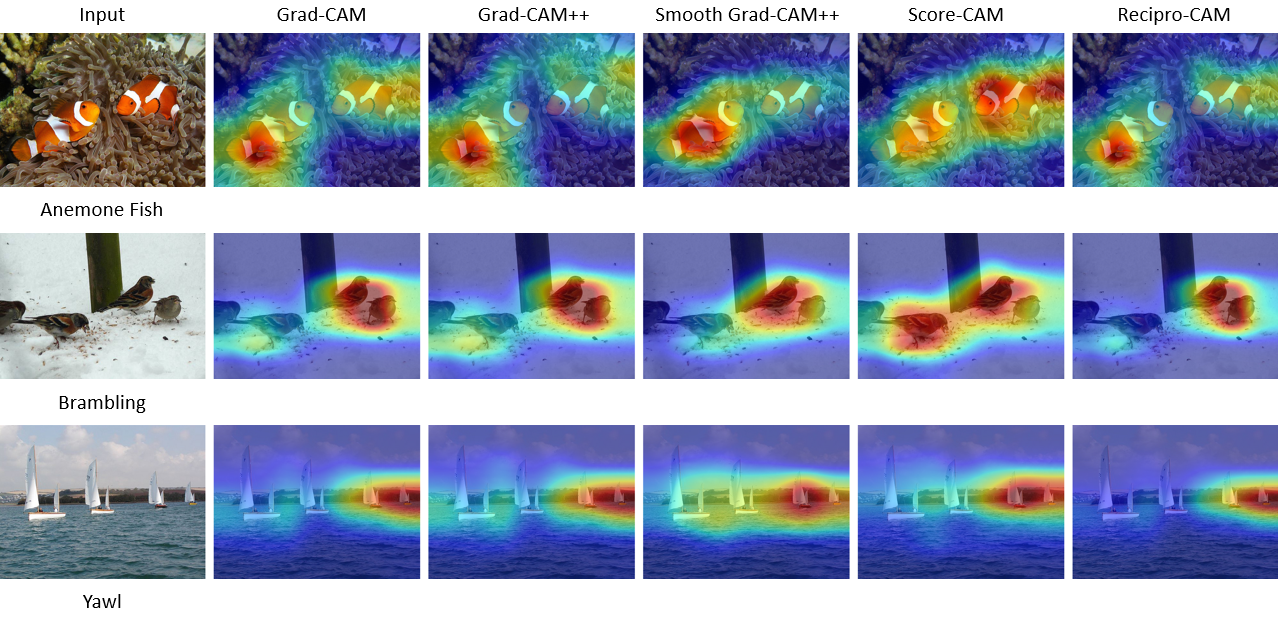}
\end{center}
\vspace{-3mm}
\caption{Same multiple objects saliency map results for Anemone Fish, Brambling, and Yawl inputs processed with Grad-CAM, Grad-CAM++, Smooth Grad-CAM++, Score-CAM, and Recipro-CAM to generate saliency maps.}
\label{fig:ex2}
\vspace{-2mm}
\end{figure}

For evaluating the localization capability of the methods, we designed the second case, and the results are presented in Figure~\ref{fig:ex2}. In this case, Recipro-CAM, Grad-CAM++, and Grad-CAM exhibited relatively separated saliency maps for the anemone fish, while Score-CAM generated a connected saliency map with a strong activation score for the coral.
Smooth Grad-CAM produced a slightly connected saliency map, but with a low saliency score for the coral. Notably, Score-CAM had the maximum saliency score at the right fish, while other CAMs had the maximum point at the left fish. 
In the brambling case, all CAMs produced different saliency map patterns. Recipro-CAM and Smooth Grad-CAM++ generated relatively low saliency scores for the left birds. Grad-CAM and Grad-CAM++ generated similar saliency maps, but Grad-CAM++ had a higher saliency score for the left birds than Grad-CAM. On the other hand, Score-CAM gave high saliency scores for all birds. Therefore, in this case, Score-CAM generated the best saliency map.
For the last yawl case, Recipro-CAM, Grad-CAM++, and Grad-CAM generated similar saliency maps, showing the highest saliency score at the fourth yawl from the left. However, Smooth Grad-CAM++ and Score-CAM showed the highest value at the third yawl from the left. In this case, Smooth Grad-CAM++ generated the best saliency map, as it covered all the yawls.

\begin{figure}[t]
\begin{center}
\includegraphics[width=6.5in]{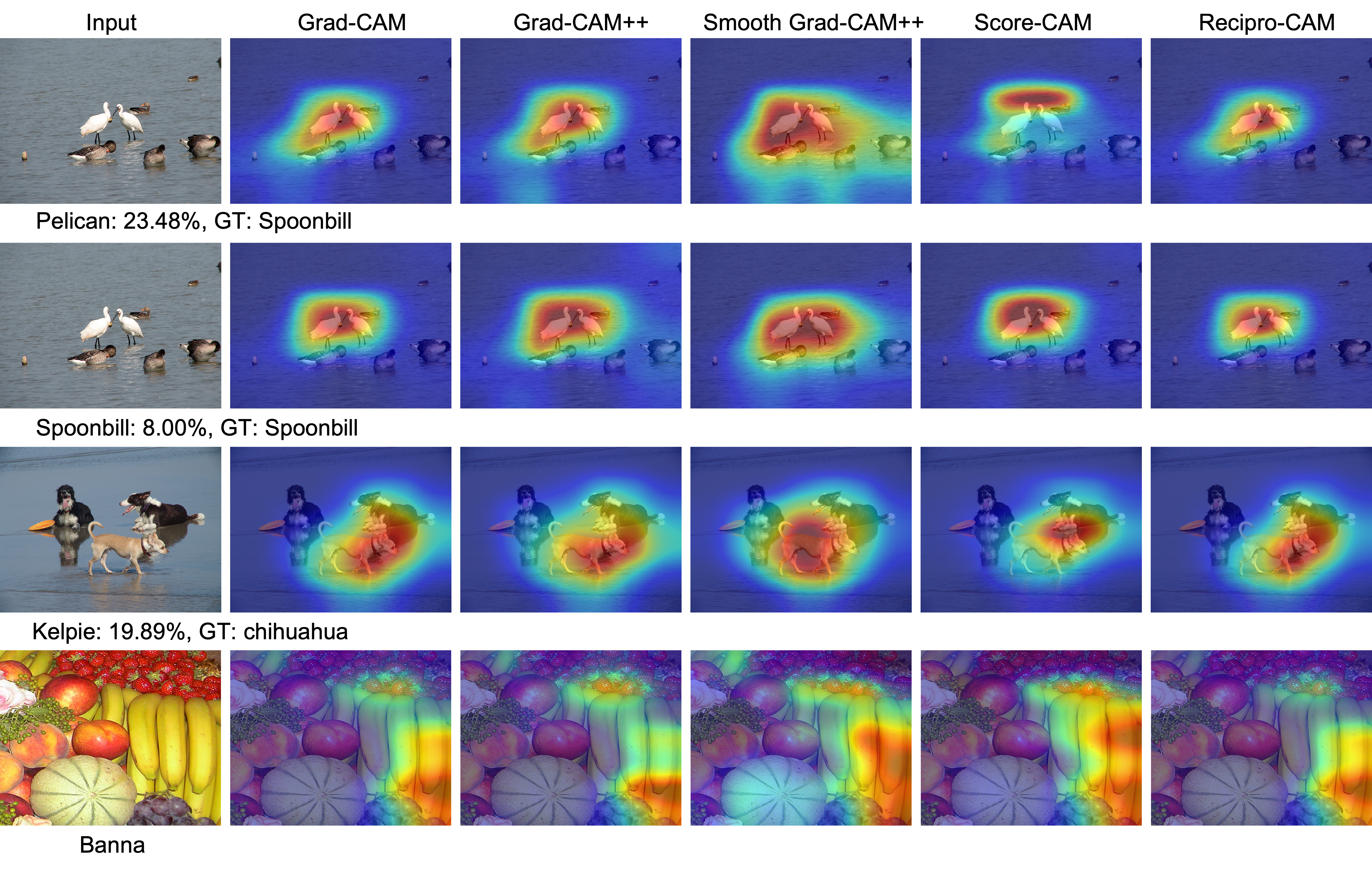}
\end{center}
\vspace{-3mm}
\caption{Results of saliency maps for multiple objects, including Spoonbill, Kelpie, and Banana inputs processed with Grad-CAM, Grad-CAM++, Smooth Grad-CAM++, Score-CAM, and Recipro-CAM. Notably, Smooth Grad-CAM++ and Score-CAM produced unusual results in this analysis.}
\label{fig:ex3}
\vspace{-2mm}
\end{figure}

The "multiple objects of different types" case was designed to evaluate the methods' resolution capability between different objects. This is because the saliency map should show class-dependent results, and different class objects should be deactivated in the image. The results are presented in Figure~\ref{fig:ex3}. 

The first row of Figure~\ref{fig:ex3} shows the result of an image containing a spoonbill and mallard objects. However, ResNet-50 predicts it as a pelican with a probability of $23.48 \%$. Recipro-CAM, Grad-CAM++, and Grad-CAM show that the network's error is related to the over-focused beak part. Smooth Grad-CAM++, on the other hand, shows the highest saliency score at the left spoonbill body part and covers the mallard objects as well. Therefore, in this case, it failed to separate different objects from the target class objects. Score-CAM, on the other hand, shows the hottest area at the lake surface, which is incorrect information.

To investigate this further, we generated a spoonbill saliency map, and the results are shown in the second row. All methods showed similar broad coverage on the body parts, but Smooth Grad-CAM++ also covered the mallard objects. 
The third row also depicts a case where ResNet-50 predicted a Kelpie with a $19.89 \%$ probability, but its ground truth is Chihuahua. All methods, except for Smooth Grad-CAM++, show a saliency map in between the second and third dogs. This suggests that ResNet-50 might have been confused by the mixed features since Kelpie has features of both dogs. Smooth Grad-CAM++, on the other hand, shows a broad saliency map covering all dogs, but the hottest area is on the second dog, which is the ground truth. Therefore, it cannot be used as a good debugging tool.
The fourth row depicts the saliency map of a banana, and all CAM methods exhibit different patterns, but they cover the banana area well, except for the Smooth Grad-CAM++ method. Its saliency map also covers the pumpkins.


In the overall analysis of multiple objects, the Score-CAM method showed unusual results. Thus, we need to check its resolution capability with other input as shown in Figure~\ref{fig:main}.
Figure~\ref{fig:main} displays the saliency maps of Recipro-CAM, Grad-CAM++, and Score-CAM, and ResNet-50 network predicts it as an aircraft carrier, but its ground truth is warplane. Recipro-CAM and Grad-CAM++ highlight the right plane and green field, so we can assume that the network might be confused with the green field because the aircraft carrier has planes on its deck, and in this case, the overlapped right plane and green field can provide similar features to the network. However, Score-CAM highlights two planes and even assigns some saliency scores to the statue of liberty. Therefore, we cannot identify the prediction error with this saliency map.


\begin{table}[]
\caption{Comparison of execution time among five methods. The execution time was measured with 1,000 inputs, and the average time was calculated, so the reported time is the execution time for a single image.}
\footnotesize \centering
\begin{tabular}{llll} \hline
& \begin{tabular}[c]{@{}l@{}}Execution\\ Time (ms) \end{tabular} & FPS & Ratio   \\ \hline
\textbf{Recipro-CAM} & 13.8 & 72.46 & - \\ 
Grad-CAM & 16.0 & 62.50 & 1.16$\times$ \\
Grad-CAM++ & 16.2 & 62.73 & 1.17$\times$ \\
SGrad-CAM++ & 77.3 & 12.94 & 5.60$\times$ \\
Score-CAM   & 2039.7 & 0.49 & 147.80$\times$  \\ \hline
\end{tabular} \label{tb:speed}
\vspace{-3mm}
\end{table}

\subsection{Performance Analysis}\label{sec:perf}
To our knowledge, previous CAM-based (white-box approach) XAI research did not consider execution performance as CAM or gradient-based CAM methods were typically fast enough, and performance was not considered an issue. However, newer algorithms such as Score-CAM and Ablation-CAM have performance issues, which are dependent on the number of feature channels, input resolution, and network capacity. Therefore, execution performance should be considered as one of the metrics for evaluating an XAI method.

We conducted experiments to measure the execution performance of Grad-CAM, Grad-CAM++, Smooth Grad-CAM++, Score-CAM, and Recipro-CAM using the Torchvision ResNet-50 ImageNet pretrained model. We used $1000$ randomly selected images from the ILSVRC2012 validation dataset, with the same image preprocessing and crop size as the quantitative analysis process. The devices used were a single Nvidia Geforce RTX 3090 and an i9-11900 Intel CPU.

Table~\ref{tb:speed} shows that Recipro-CAM demonstrated the best execution performance (Time: $13.8$ ms, FPS: $72.46$), followed by Grad-CAM and Grad-CAM++ with similar performance to Recipro-CAM. However, Smooth Grad-CAM++ exhibited relatively lower performance (Time: $77.3$ ms, FPS: $12.94$).
On the other hand, Score-CAM exhibited the worst performance (Time: $2.0397$ s, FPS: $0.49$), which is approximately $148 \times$ slower than Recipro-CAM. Hence, Score-CAM has a weak point in its performance.   

\section{Conclusion}
The Recipro-CAM method is a novel approach to generate saliency maps for CNNs that does not require gradient information. The approach involves masking the extracted feature maps to leverage the correlation between the activation maps and the network output. Our experiments demonstrate that Recipro-CAM outperforms other saliency map generation methods in terms of both execution time and performance, achieving state-of-the-art results on the ADCC metric. Notably, Recipro-CAM is approximately $148 \times$ faster than Score-CAM, a previously proposed gradient-free method. Furthermore, Recipro-CAM's execution time is faster than that of gradient-based methods. 

\bibliographystyle{unsrtnat}
\bibliography{references}  






\end{document}